\RequirePackage{fix-cm}
\documentclass[smallextended]{svjour3}       
\smartqed  
\usepackage{amssymb}
\usepackage{graphicx}
\usepackage{algorithm, algorithmic}
\usepackage{amsmath}
\usepackage{color}
\usepackage{multirow}
\usepackage[table,xcdraw]{xcolor}

\begin{document}
	
	\title{Automated Surgical Skill Assessment in RMIS Training}
	
	
	\author{Aneeq Zia \and
		Irfan Essa 
	}
	
	
	\institute{A. Zia \\
		\email{aneeqzia@gmail.com}\\
		\\
		College of Computing, Georgia Institute of Technology \\
		Atlanta, GA, USA 30332 \\
	}
	
	\date{Received: date / Accepted: date}

	\maketitle
\maketitle

\begin{abstract}

\textit{Purpose}: 
Manual feedback in basic RMIS training can consume a significant amount of time from expert surgeons' schedule and is prone to subjectivity. While VR-based training tasks can generate automated score reports, there is no mechanism of generating automated feedback for surgeons performing basic surgical tasks in RMIS training. In this paper, we explore the usage of different holistic features for automated skill assessment using only robot kinematic data and propose a weighted feature fusion technique for improving score prediction performance. Moreover, we also propose a method for generating \textit{`task highlights'} which can give surgeons a more directed feedback regarding which segments had the most effect on the final skill score.

\noindent \textit{Methods}:
We perform our experiments on the publicly available JIGSAWS dataset and evaluate four different types of holistic features from robot kinematic data - Sequential Motion Texture (SMT), Discrete Fourier Transform (DFT), Discrete Cosine Transform (DCT) and Approximate Entropy (ApEn). The features are then used for skill classification and exact skill score prediction. Along with using these features individually, we also evaluate the performance using our proposed weighted combination technique. The task highlights are produced using DCT features.

\noindent \textit{Results}:
Our results demonstrate that these holistic features outperform all previous HMM based state-of-the-art methods for skill classification on the JIGSAWS dataset. Also, our proposed feature fusion strategy significantly improves performance for skill score predictions achieving up to 0.61 average spearman correlation coefficient. Moreover, we provide an analysis on how the proposed task highlights can relate to different surgical gestures within a task.

\noindent \textit{Conclusions}: 
Holistic features capturing global information from robot kinematic data can successfully be used for evaluating surgeon skill in basic surgical tasks on the da Vinci robot. Using the framework presented can potentially allow for real time score feedback in RMIS training and help surgical trainees have more focused training.

\keywords{Robot-assisted surgery \and Surgical skill assessment \and Feature fusion}

\end{abstract}

\section{Introduction}

With the rapidly increasing amount of Robot-Assisted Minimally Invasive Surgery (RMIS) around the world, the focus on robotic surgical training has increased tremendously. Typical robotic surgery training includes simulator based and dry lab exercises like suturing, knot tying and needle passing. Training on these tasks is crucial since it forms the base for advanced training procedures on pigs, cadavers and eventually, humans. However, the current assessment on such dry lab exercises is done manually by supervising surgeons which makes it prone to subjectivity and reduces the overall efficiency of training. In order to reduce subjectivity, many medical schools are starting to adopt Objective Structured Assessment of Technical Skills (OSATS) as a grading system \cite{28martin1997objective}. OSATS consists of different grading criteria like Respect for Tissue (RT), Time and Motion (TM), Flow of Operation (FO), Overall Performance (OP) and Quality of Final Product (QP). However, this grading is still done manually making it extremely time consuming. 

Much of the literature in basic RMIS training has focused on developing methods for recognizing surgical gestures \cite{reiley2009decomposition,haro2012surgical,dipietro2016recognizing,ahmidi2017dataset}. Although recognizing surgical gestures within a task can be helpful for skill assessment, treating the data from tasks as a whole reduces the complexity of the problem and has been shown to work well enough \cite{zia2015automated,zia2016automated,zia2017video,sharmavideo}. Some of the recent approaches for automated surgical skills assessment in RMIS training have tried to use variants of HMM \cite{tao2012sparse} given data from a task. While HMM's can be effective in modeling temporal data, we hypothesize that extracting features capturing global information from time series data can be more indicative of surgeon skill. Moreover, to the best of our knowledge, all the works in the surgical skills assessment domain have only proposed methods for predicting overall scores of a task. However, although that is the important first step towards better feedback, we feel that it is extremely important to give surgeons more directed feedback as to which part of a particular task contributed to their high or low score. In this paper, we present a detailed analysis on skill assessment for basic RMIS training and list our main contributions below. 

\emph{\textbf{Contributions:}} (1) We propose a framework for automated surgical skills assessment in RMIS training, and show that texture, frequency and entropy based features outperform all previous HMM based state-of-the-art techniques on JIGSAWS dataset using kinematic data. (2) We propose a weighted feature fusion technique for skill score prediction. (3) We provide a detailed analysis on skill assessment on JIGSAWS dataset and show the role played by 
different features in score predictions. (4) We propose a technique for generating task highlights that can provide surgeons with more directed feedback as to which parts of a task had the most positive/negative impact on the final score prediction.

\section{Background}

Automated surgical skills assessment has been of interest to researchers for a long time. Recent works have shown promising results in both RMIS and video based basic surgical skills assessment.

In video based approaches, most works employ Spatio-Temporal Interest Points (STIP) \cite{29laptev2005space} to capture motion information and use them to develop models for skill prediction \cite{sharma2014video,zia2015automated,zia2016automated,zia2017video,bettadapura2013augmenting,sharmavideo}. \cite{sharmavideo} proposed Sequential Motion Texture (SMT) that used texture features of frame kernel matrices for skill prediction. In \cite{zia2015automated,zia2016automated}, the authors used the repeatability in motions via frequency features (DCT and DFT) to classify surgeon skill level. More recently, \cite{zia2017video} proposed to encode the predictability in surgical motions using approximate entropy (ApEn) and cross approximate entropy (XApEn) for skill assessment. In the computer vision literature, frequency and entropy based features have been shown to perform good for sports quality assessment as well \cite{pirsiavash2014assessing,venkataraman2015dynamical}

For assessment of surgical skills in RMIS, one of the earlier works proposed a variant of HMM - sparse HMM \cite{tao2012sparse}. Other works like \cite{nisky2015teleoperated} studied the differences in needle-driving movements and reported significant differences between beginner and expert surgeons. In \cite{ahmidi2013string}, the authors proposed descriptive curve coding-common string model (DCC-CSM) for simultaneous surgical gesture recognition and skill assessment. \cite{fard2016machine} used SVM on basic metrics like time for completion, path length, speed etc, for skill evaluation. More recently, some works have explored the use of crowd sourcing techniques to evaluate surgeon skill \cite{ershad2016meaningful}.

Although the previous works have shown promising results on RMIS based skill prediction, none of them explored the usage of features capturing the repeatability and predictability in surgical motions (like frequency \cite{zia2015automated,zia2016automated} and entropy based \cite{zia2017video}) from robot kinematic data. We hypothesize that such features would be able to capture more skill relevant information since expert robotic surgeons tend to have smoother and predictable motions as compared to beginners. Moreover, inspired by the work in \cite{pirsiavash2014assessing}, we hypothesize that frequency based features can be used to evaluate the impact any short segment has on the final score prediction for surgical tasks using inverse transforms.

\section{Methodology}
\subsection{Skill Classification/Score Prediction}

\begin{figure}[t]
	\centering
	\includegraphics[width=1.0\columnwidth]{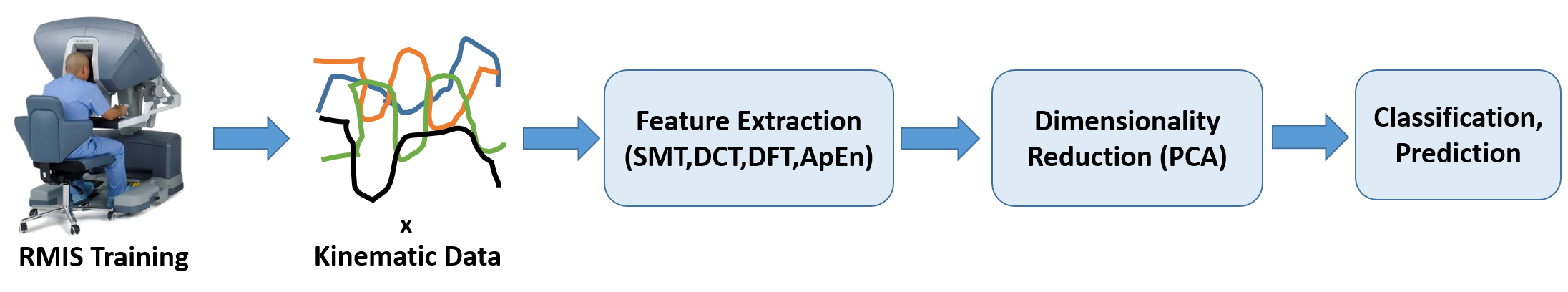}
	\caption{Flow diagram of the proposed framework for robotic surgical skills assessment.}
	\label{fig:flow_diagram}

\end{figure}
As opposed to previous proposed works on using different variants of HMMs for skill assessment, we evaluate holistic features for predicting skill level using robot kinematics data. Figure \ref{fig:flow_diagram} shows the proposed pipeline. For a given $D$-dimensional time series $S \in \Re^{D \times L}$, where $L$ is the number of frames, we extract 4 different types of features: Sequential Motion Texture (SMT), Discrete Fourier Transform (DFT), Discrete Cosine Transform (DCT) and Approximate Entropy (ApEn). The dimensionality of the features is reduced using Principal Component Analysis (PCA) before classification/prediction. We give details of the feature types, fusion method and the prediction model below.

\noindent\textbf{SMT}: 
Sequential motion texture was implemented as presented in the original paper \cite{sharmavideo}. The time series is divided into $N_w$ number of windows. A frame kernel matrix is calculated after which Gray Level Co-Occurence Matrices (GLCM) texture features (20 in total) are evaluated resulting in a feature vector $\phi_{SMT} \in \Re^{20N_w}$.

\noindent\textbf{DCT/DFT}:
Frequency based features have proven to work well for video based assessment of actions like olympic sports \cite{pirsiavash2014assessing} and basic surgical tasks \cite{zia2015automated,zia2016automated}. We evaluate DCT and DFT coefficients for each dimension of the robot kinematics time series. This results in a matrix of frequency components $F \in \Re^{D \times L}$. The lowest $Q$ components from each dimension are then concatenated together to make the final feature vector $\phi_{DCT/DFT} \in \Re^{DQ}$. Using low frequency features would eliminate any high frequency noise that could have resulted during data capture.

\noindent\textbf{ApEn}: Expert surgeons tend to have a more fluent and predictable motion as compared to beginners. Therefore, a measure of predictability in temporal kinematic data can potentially help differentiate between varying skill levels. Approximate entropy is a measure of predictability in a time series data \cite{pincus1991approximate} and has been used in recent literature for activity assessment \cite{venkataraman2015dynamical,zia2017video}. We extract ApEn features from our robot kinematic time series data as presented in \cite{zia2017video}. 
Evaluating ApEn for all dimensions of the time series data results in a feature vector $\phi_{ApEn} \in \Re^{DR}$, where $R$ is the number of radius values used in evaluation per dimension.

\begin{figure}[t!]
	\centering
	\includegraphics[width=1.0\columnwidth]{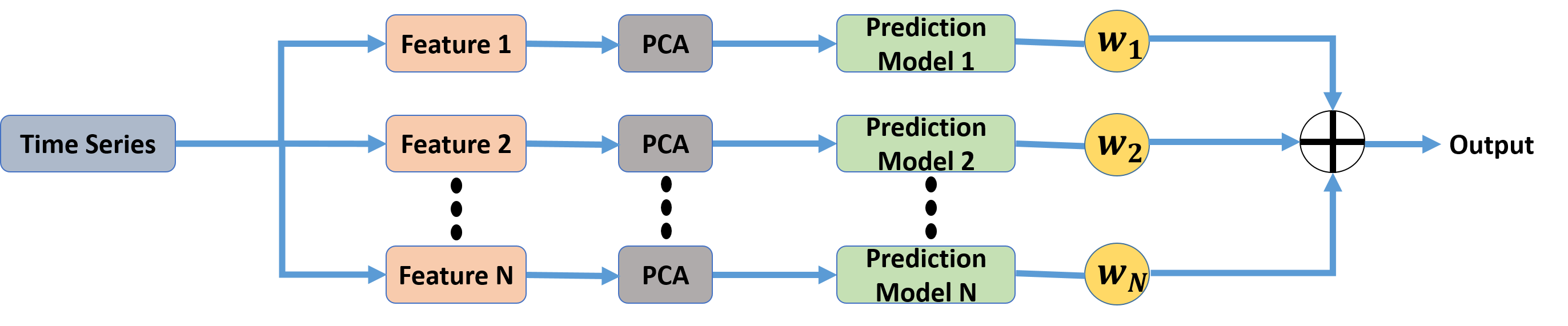}
	\caption{Weighted feature fusion for modified-OSATS score and GRS prediction.}
	\label{fig:fusion}
\end{figure}

\noindent\textbf{Feature Fusion}: We propose a weighted feature fusion technique for skill prediction (as shown in Figure \ref{fig:fusion}). The outputs of different prediction models are combined to produce a skill score. We take our training time series data and evaluate each feature type to produce a training feature matrix $\phi_f \in \Re^{n \times D}$, where $f$ corresponds to a the feature type used, $n$ is the number of training samples and $D$ is the dimensionality of the feature type. The output $y_f \in \Re^{n}$ corresponding to each $\phi_f$ is then evaluated using the prediction model. A matrix of outputs from different features $Y \in \Re^{n \times F}$ is generated by concatenating all the $y_f$, where $F$ corresponds to total number of features used. Given the ground truth predictions $G \in \Re^{n}$, the optimal weights vector $w^* \in \Re^{F}$ is then evaluated by solving a simple least squares as $w^* = \underset{w}{\mathrm{argmin}}  ||Yw - G||^2_2$. For a given test set, the output $\hat{y_{test}}$ is then calculated using $\hat{y_{test}} = Y_{test}w^*$.

\noindent\textbf{Classification/Prediction}: We use a simple nearest neighbor classifier for classification of skill levels. For exact score prediction, we use a linear support vector regression (SVR) model \cite{SVR}.

\subsection{Task Highlights}

Apart from giving feedback to surgeons in terms of skill score predictions, it could be of great help to surgeons if they knew which parts of the task impacted their final score the most. This can potentially allow surgeons to focus more on specific gestures that contribute to low scores. We define the impact of a segment as the amount by which the predicted score would change if that segment was not observed. In order to do this, we need to evaluate the inferred feature vector had we not observed a particular segment of the data. 

Few works have presented approaches for measuring impact of different segments on overall skill score predictions. For example, in \cite{pirsiavash2014assessing}, the authors presented a frequency features based approach using human pose for evaluating the impact of a particular segment on final score prediction in Olympic sports. Similar to their work, we present a DCT feature based approach for generating \textit{`task highlights'} using robot kinematics data. For a given d-th dimension of the kinematic time series $S(d) \in \Re^{L}$, the corresponding DCT features $F(d)\in \Re^{L}$ are evaluated using $F(d) = A S(d)$, where $A \in \Re^{L \times L}$ is the DCT transformation matrix. Taking $B=A^+$ as the inverse cosine transformation matrix (where $A^+$ denotes the pseudo-inverse of $A$), the DCT equation can be written as $F(d) = B^+ S(d)$. Now, if the data from frames $n_1$ till $n_2$ were to be removed, we can evaluate the inferred DCT feature vector by $\hat{F}(d) = (B_{n_1:n_2})^+ S(d)$, where $B_{n_1:n_2})$ is the matrix $B$ with rows $n_1$ till $n_2$ removed. $\hat{F}$ will essentially have inferred the missing segment by the most likely kinematics signal given the frequency spectrum of the rest of the signal. Since $\hat{F}$ will have the same dimensionality as $F$, we can use the same SVR model for score prediction. The final impact of the segment on skill score is then evaluated by $impact = \psi - \hat{\psi}$, where $\psi$ is the predicted score using whole sequence and $\hat{\psi}$ is the inferred score with a missing segment. The surgical task highlights can be generated by evaluating the $impact$ on a running window.


\begin{figure}[t!]
	\centering
	\includegraphics[width=1.0\columnwidth]{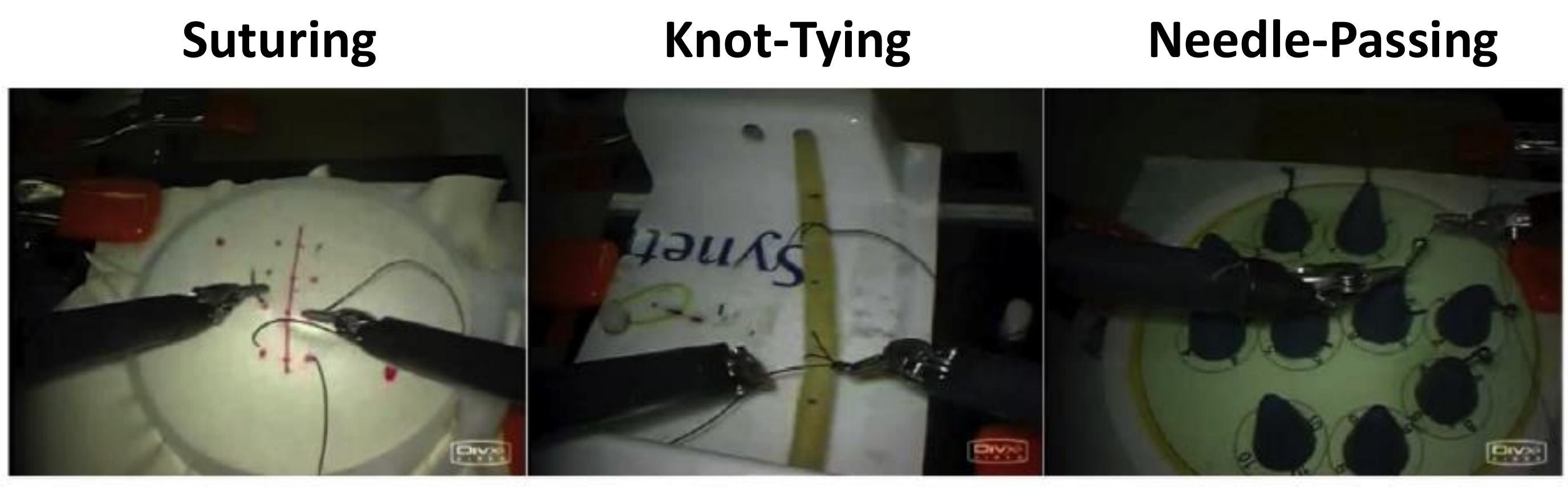}
	\caption{Sample frames from the 3 tasks in the JIGSAWS dataset \cite{gao2014jhu}.}
	\label{fig:sample_frames}
\end{figure}

\section{Experimental Evaluation}
\textbf{Dataset:} Our proposed framework is evaluated on the publicly available JIGSAWS dataset \cite{gao2014jhu}. This dataset consists of kinematics and video data from 8 participants for three robotic surgical tasks: Suturing, Knot Tying and Needle Passing. Figure \ref{fig:sample_frames} shows sample frames for each task. We only use kinematic data for our analysis and employ the standard LOSO (\textit{leave-one-supertrial-out}) and LOUO (\textit{leave-one-user-out}) cross validation setups. For LOSO, we leave one randomly selected trial from each surgeon out for testing and repeat this 20 times. For LOUO, we leave all trials from one surgeon out for testing. The dataset has ground truth skill labels of three categories: self-proclaimed, modified-OSATS and global gating score (GRS). Self-proclaimed category has three skill levels (dependent on the amount of hours spent on the system) $-$ novice ($<10$ hrs), intermediate ($10-100$ hrs) and expert ($>100$ hrs). The modified-OSATS scores are based on six criteria on a scale of 1-5 and are generated by an expert watching the videos while grading them. This is different from the original OSATS \cite{28martin1997objective} (as described in introduction section) since it contains an extra criteria of suture handling (SH) and that none of the criteria are graded as Pass/Fail. The GRS is a sum of all individual modified-OSATS scores.

\noindent\textbf{Parameter estimation:} We use the original feature implementations as presented in \cite{sharmavideo,zia2015automated,zia2017video}. In SMT, we use number of windows $N_w=10$ and evaluate Gray Level Co-Occurence Matrices (GLCM) texture features with 8 gray levels resulting in a 200-dimensional feature vector. For frequency features, we take the lowest 50 components ($Q = 50$) for each dimension of the time series and concatenate them resulting in a $50D$-dimensional feature vector, where $D$ is the dimension of time series (76 in our case). In calculating approximate entropy ($ApEn$), we use radius $r = [0.1, 0.13, 0.16, 0.19, 0.22, 0.25]$ resulting in a $6D$-dimensional feature vector. A value of $1$ was used for both $m$ and $\tau$.

We use Principal Component Analysis (PCA) for dimensionality reduction before passing features onto the classifier or the regression model. This was done since a lower performance was observed using original feature dimensionality. In order to estimate the optimal number of PCA componenets $D_{PCA}$, we evaluate performance for $D_{PCA}$ ranging from 10 to 3000 for all tasks for each feature type. The value of $D_{PCA}$ corresponding to highest average performance accross all tasks was selected. For score predictions, we need to estimate an optimal value for the regularization parameter $C$ in SVR. For each feature type, we evaluated the average correlation coefficient (over all modified-OSATS) for $C \in [10^{-7}, 10^{-6}, \ldots , 10^{6} , 10^{7}]$ and selected the best performing value of C for evaluations. The optimal values of $D_{PCA}$ and $C$ are given in Table \ref{Parameters}. Please note that all parameters were strictly tuned on the training data only for both validation setups. This includes the weights being estimated for the fusion of different prediction models.

For task highlights generation, we use 50 lowest DCT features (same as for classification/prediction) with a running window of length 100.

\begin{table}[t!]
	\centering
	\caption{Table showing optimal number of PCA components estimated. For prediction, the optimal value of the regularization parameter C is given within parantheses.}
	\label{Parameters}
	\begin{tabular}{|c|c|c|c|c|}
		\hline
		& SMT        & DCT           & DFT          & ApEn       \\ \hline
		Classification & 50         & 150           & 150          & 40         \\ \hline
		Prediction     & 10 ($10^2$) & 1000 ($10^{-6}$) & 250 ($10^{-6}$) & 40 ($10^4$) \\ \hline
	\end{tabular}
\end{table}

%


\section{Results and Discussion}

We evaluate the proposed features for skill classification and modified-OSATS based score prediction using the JIGSAWS dataset. For classification, we compare the performance of these features with previous HMM based state-of-the-art methods \cite{tao2012sparse}. Table \ref{Classification_Results}  shows results for self proclaimed skill level classification in the JIGSAWS dataset. As evident, using holistic features significantly out-perform previous approaches of using different variants of HMMs. Specifically, ApEn performs significantly better than all other methods. This is interesting to note since experts (with $>100$ hrs of practice) would have smoother motions as compared to beginners (with $<10$ hrs of practice) making their movements more `predictable', and hence easily differentiated using ApEn features.
\begin{table}[t!]
	\centering
	\caption{Self proclaimed skill classification results}
	\label{Classification_Results}
	\begin{tabular}{|c|c|c|c|c|c|c|}
		\hline
		\multirow{2}{*}{} & \multicolumn{2}{c|}{Suturing}    & \multicolumn{2}{c|}{Knot Tying}   & \multicolumn{2}{c|}{Needle Passing} \\ \cline{2-7} 
		& LOSO           & LOUO            & LOSO            & LOUO            & LOSO             & LOUO             \\ \hline
		MFA-HMM           & 92.3           & 38.5            & 86.1            & 44.4            & 76.9             & 46.2             \\ \hline
		KSVD-HMM          & 97.4           & 59              & 94.4            & 58.3            & 96.2             & 26.9             \\ \hline
		SMT               & 99.70          & 35.3            & 99.6            & 32.3            & 99.9             & 57.1             \\ \hline
		DCT               & \textbf{100} & 64.7            & 99.7            & 54.8            & 99.9             & 35.7             \\ \hline
		DFT               & \textbf{100} & 64.7            & \textbf{99.9} & 51.6            & 99.9             & 46.4             \\ \hline
		ApEn              & \textbf{100} & \textbf{88.2} & \textbf{99.9} & \textbf{77.4} & \textbf{100}   & \textbf{85.7}  \\ \hline
	\end{tabular}
\end{table}

\begin{table}[]
	\centering
	\caption{OSATS scores and GRS prediction results. Each cell contains two numbers in the form $\rho_{OSATS} ~\vert~ \rho_{GRS}$, where the first number is the value of $\rho$ averaged over all OSATS and the latter is the value of $\rho$ for GRS prediction. ``*" means a $p-$value $<0.05$ for the corresponding $\rho$.}
	\label{Prediction_Results}
	\resizebox{\textwidth}{!}{%
	\begin{tabular}{|c|c|c|c|c|c|c|}
		\hline
		\multirow{2}{*}{} & \multicolumn{2}{c|}{Suturing}                                                        & \multicolumn{2}{c|}{Knot Tying}                                                      & \multicolumn{2}{c|}{Needle Passing}                                             \\ \cline{2-7} 
		& LOSO                                      & LOUO                                     & LOSO                                 & LOUO                                          & LOSO                                 & LOUO                                     \\ \hline
		SMT               & ~0.25 $\vert$ 0.46*                  & -0.08 $\vert$ -0.28                      & 0.41* $\vert$ 0.39*                  & 0.18 $\vert$ 0.21                             & -0.12 $\vert$ 0.09                   & ~~~0.07 $\vert$ -0.60*      \\ \hline
		DCT               & 0.57* $\vert$ 0.68*                       & 0.10 $\vert$ 0.08                        & ~0.59* $\vert$ \textbf{0.76*} & ~0.49 $\vert$ 0.73*                      & ~~0.22 $\vert$ 0.26*        & -0.16 $\vert$ 0.09                       \\ \hline
		DFT               & 0.45* $\vert$ 0.49*                       & -0.28 $\vert$ -0.29                      & ~0.31 $\vert$ 0.32*             & 0.46* $\vert$ 0.47*                           & ~0.44* $\vert$ \textbf{0.53*} & ~0.37 $\vert$ 0.19                  \\ \hline
		ApEn              & 0.31* $\vert$ 0.49*                       & ~0.43 $\vert$ 0.40*                 & ~0.26 $\vert$ 0.14*             & 0.02 $\vert$ 0.12                             & 0.16 $\vert$ 0.06                    & ~~0.21 $\vert$ -0.21            \\ \hline
		SMT+DCT           & 0.48* $\vert$ 0.61*                       & 0.01 $\vert$ 0.01                        & \textbf{0.66*}$\vert$ 0.71*        & ~~~0.46 $\vert$ \textbf{0.78*} & ~0.14 $\vert$ -0.16             & ~-0.23 $\vert$ -0.14                \\ \hline
		SMT+DFT           & 0.40* $\vert$ 0.60*                       & ~-0.21 $\vert$ -0.49*               & ~0.36 $\vert$ 0.39*             & 0.52* $\vert$ 0.48*                           & 0.39* $\vert$ 0.54*                  & ~0.33 $\vert$ 0.13                  \\ \hline
		SMT+ApEn          & 0.28* $\vert$ 0.35*                       & ~~0.41 $\vert$ \textbf{0.42*} & ~0.18 $\vert$ 0.36*             & 0.06 $\vert$ 0.12                             & ~0.12 $\vert$ -0.06             & ~~0.15 $\vert$ -0.29            \\ \hline
		SMT+DCT+DFT       & 0.57* $\vert$ 0.64*                       & 0.16 $\vert$ 0.10                        & 0.58* $\vert$ 0.70*                  & \textbf{0.56*}$\vert$ 0.73*                 & 0.36* $\vert$ 0.38*                  & 0.50* $\vert$ 0.23                       \\ \hline
		DCT+DFT           & 0.56* $\vert$ 0.66*                       & 0.13 $\vert$ 0.14                        & 0.53* $\vert$ 0.68*                  & 0.55* $\vert$ 0.73*                           & 0.41* $\vert$ 0.47*                  & \textbf{0.53*} $\vert$ \textbf{0.28} \\ \hline
		DCT+DFT+ApEn      & \textbf{0.59*} $\vert$ \textbf{0.75*} & 0.43* $\vert$ 0.37*                      & 0.57* $\vert$ 0.63*                  & ~0.48 $\vert$ 0.60*                      & ~~0.37 $\vert$ 0.46*        & ~~0.23 $\vert$ 0.25             \\ \hline
		SMT+DCT+DFT+ApEn  & 0.47* $\vert$ 0.66*                       & \textbf{0.45*}$\vert$ 0.37*            & 0.55* $\vert$ 0.61*                  & ~0.49 $\vert$ 0.62*                      & \textbf{0.45*}$\vert$ 0.45*        & ~~-0.21 $\vert$ -0.19           \\ \hline
	\end{tabular}
	}
\end{table}

\begin{table}[t!]
	\centering
	\caption{ Values of $\rho$ averaged over all three tasks for the corresponding feature types in the form  $\rho_{OSATS} ~\vert~ \rho_{GRS}$.}
	\label{Average_Prediction_Results}
	\begin{tabular}{|c|c|c|}
		\hline
		& LOSO                                    & LOUO                               \\ \hline
		SMT              & 0.18 $\vert$ 0.31                       & ~0.05 $\vert$ -0.22           \\ \hline
		DCT              & 0.46 $\vert$ 0.57                       & 0.14 $\vert$ 0.24                  \\ \hline
		DFT              & 0.40 $\vert$ 0.45                       & 0.19 $\vert$ 0.12                  \\ \hline
		ApEn             & 0.24 $\vert$ 0.23                       & 0.22 $\vert$ 0.10                  \\ \hline
		SMT+DCT          & 0.43 $\vert$ 0.39                       & 0.08 $\vert$ 0.22                  \\ \hline
		SMT+DFT          & 0.38 $\vert$ 0.51                       & 0.22 $\vert$ 0.04                  \\ \hline
		SMT+ApEn         & 0.20 $\vert$ 0.22                       & 0.21 $\vert$ 0.08                  \\ \hline
		SMT+DCT+DFT      & 0.50 $\vert$ 0.57                       & \textbf{0.41}$\vert$ 0.36        \\ \hline
		DCT+DFT          & 0.50 $\vert$ 0.60                       & 0.40 $\vert$ 0.38                  \\ \hline
		DCT+DFT+ApEn     & \textbf{0.51} $\vert$ \textbf{0.61} & ~0.38 $\vert$ \textbf{0.41} \\ \hline
		SMT+DCT+DFT+ApEn & 0.49 $\vert$ 0.58                       & 0.24 $\vert$ 0.27                  \\ \hline
	\end{tabular}
\end{table}

Table \ref{Prediction_Results} shows the results for modified-OSATS and global rating score predictions. We use spearman's correlation coefficient `$\rho$' as an evaluation metric and check for statistical significance using the $p$-value. For modified-OSATS score prediction, we show the value of $\rho$ averaged over all six criteria, whereas, the GRS $\rho$ values are given as is. Feature combination results presented in Table \ref{Prediction_Results} are evaluated using weighted feature fusion as described in methodology section. Overall, we can see that individual features and their combinations achieve good results for the LOSO setup. On the other hand, we see a comparatively low performance on LOUO setup. This is because LOUO is a harder validation scheme due to less data for training phase. However, using the proposed feature combination significantly improves performance over individual features. In general, frequency features seem to perform well when used individually or in combination with other features. We can also see an overall lower performance across all features for the needle-passing task. The reason for this could be that needle-passing is a relatively less repetitive task as compared to the other two. Since the features we use try to differentiate between different skill levels using data repeatability, they perform less well for needle-passing. Table \ref{Average_Prediction_Results} shows the average of $\rho$ values over all three tasks (as given in Table \ref{Prediction_Results}) for each feature type. We observe that DCT+DFT+ApEn performs best on average for OSATS and GRS score prediction.

\begin{figure}[t!]
	\centering
	\includegraphics[width=1.0\columnwidth]{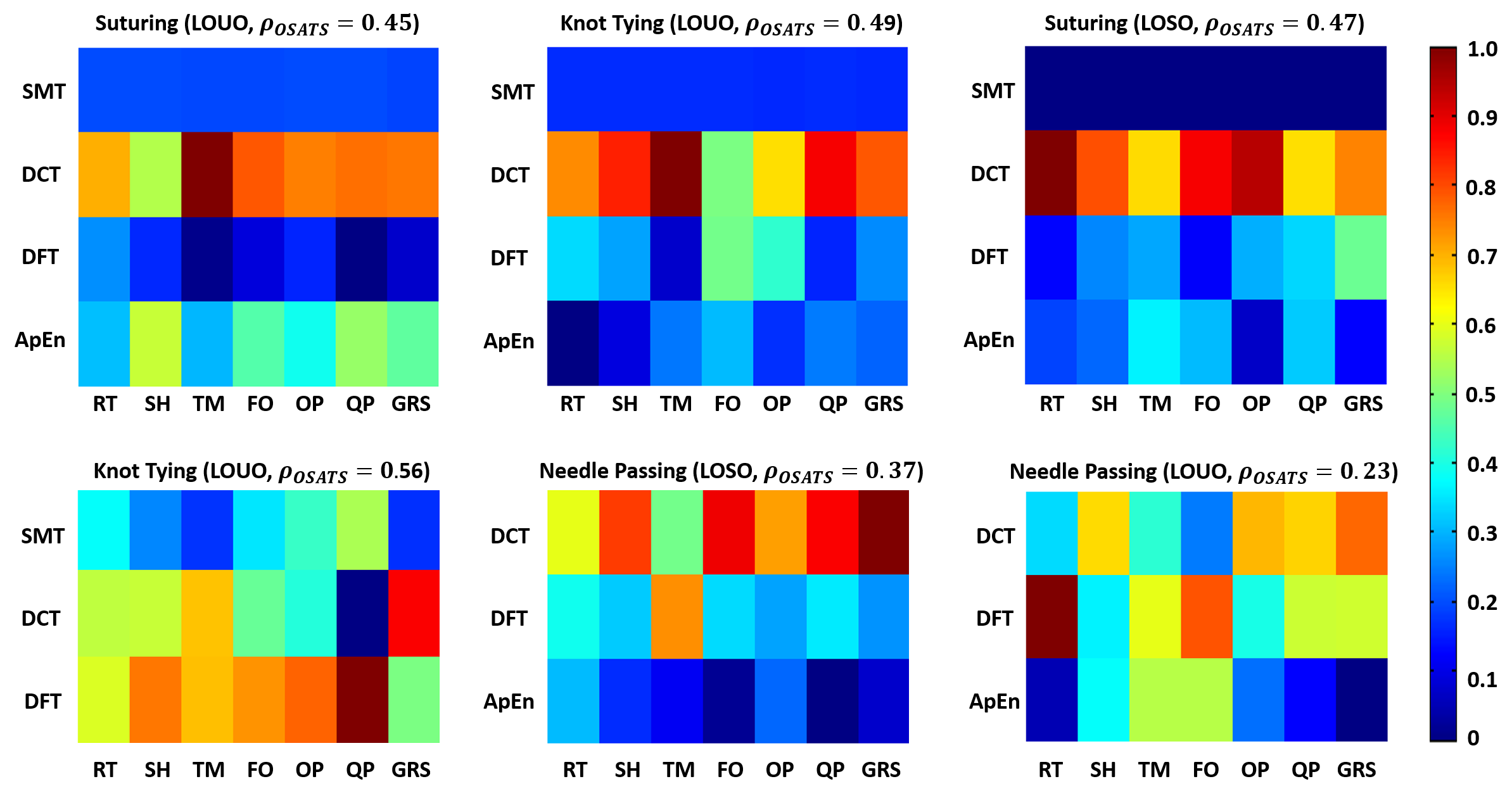}
	\caption{Heatmaps of weight assignments of different features. Each column shows the weight vector $w*$ (scaled from 0 to 1) for the corresponding OSATS criteria or GRS. For each heatmap, the features used in combination are shown next to each row and the corresponding task, validation scheme and average $\rho$ (over OSATS) are also shown. (Please view this figure in color)}
	\label{fig:heatmaps}
	
\end{figure}

In order to analyze the role of different features in the proposed weighted late fusion for skill prediction, we generate heatmaps of the weight vectors learned and show a few of them in Figure \ref{fig:heatmaps}. It can be seen that DCT features get assigned the highest weight in most of the cases. DFT and ApEn features generally have similar weight assignments whereas SMT always gets assigned a low weight. This shows that DCT features capture the most skill relevant information which is also evident from its high performance compared to other individual features in Table \ref{Prediction_Results}. 

\begin{figure}[t!]
	\centering
	\includegraphics[width=1.0\columnwidth]{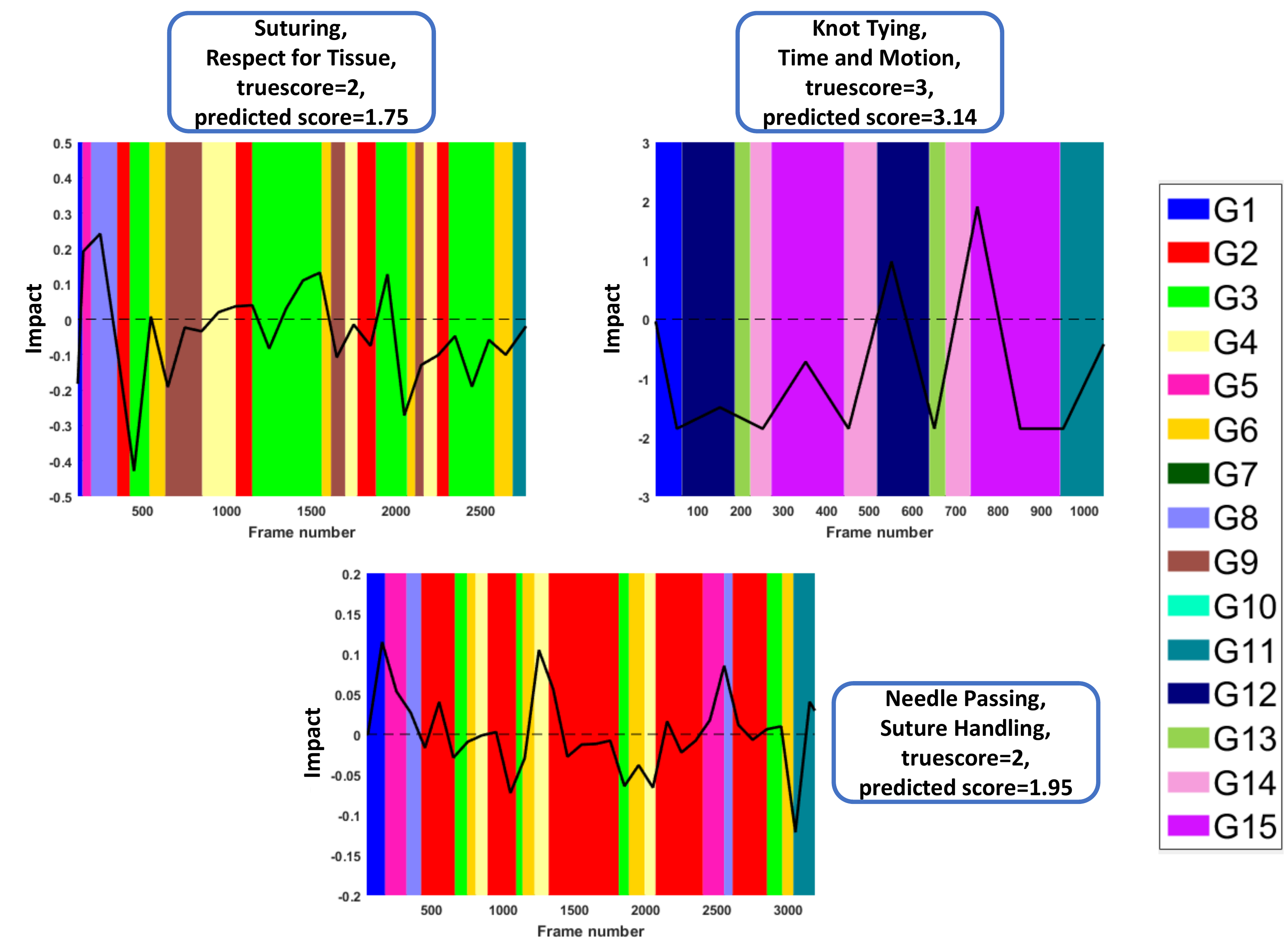}
	\caption{Sample task highlights. The y-axis on each plot corresponds to the impact (as defined in methodology section) with number of frames on the x-axis. The task type, modified-OSATS criteria, ground truth score, and the predicted score from our model using DCT features on the whole sequence, are given in boxes next to each plot. The color coding for the different gestures is also provided.}
	\label{fig:highlights}
	
\end{figure}

Figure \ref{fig:highlights} shows some sample task highlights constructed by following the procedure described in the methodology section. We overlay the impact scores plot on color coded gestures for getting better insights. The gestures used are the same as presented in the original dataset paper \cite{gao2014jhu}. For completeness, the gesture vocabulary of JIGSAWS dataset is given in Table \ref{table-gesture_vocab}.  The segments where the impact scores are negative indicate that these parts had a adverse effect on the final score, and vice versa. There are some interesting points that we can note from these plots that make intuitive sense. For example, in the suturing plot, we can observe that the impact score has maximum variations for G3 (i.e. Pushing needle through tissue). Since we predict for RT criteria in this case, one would expect that a `good' or `bad' push of a needle through the tissue should have the maximum impact on final skill score prediction. Similarly, for knot tying, we can see high positive and negative impact scores for G15 (i.e. Pulling suture with both hands). Again, this makes intuitive sense since G15 is important for knot tying task. We can draw similar insights for needle-passing (considering G2, G4 and G5) as well. Although there are no ground-truth highlights to compare our results to (and it probably would be an extremely tedious task to generate such ground-truths), we believe that such impact score plots can tremendously help surgeons in understanding the parts within a task that they need to improve on. As a result, surgical trainees can direct their time and training on specific gestures within a task which can potentially allow them to move through their learning curves much faster.

\begin{table}[t]
	\centering
	\caption{Gesture vocabulary \cite{gao2014jhu}.}
	\label{table-gesture_vocab}
	\begin{tabular}{|c|c|}
		\hline
		Gesture ID & Description                                     \\ \hline
		G1         & Reaching for needle with right hand             \\ \hline
		G2         & Positioning needle                              \\ \hline
		G3         & Pushing needle through tissue                   \\ \hline
		G4         & Transferring needle from left to right          \\ \hline
		G5         & Moving to center with needle in grip            \\ \hline
		G6         & Pulling suture with left hand                   \\ \hline
		G7         & Pulling suture with right hand                  \\ \hline
		G8         & Orienting needle                                \\ \hline
		G9         & Using right hand to help tighten suture         \\ \hline
		G10        & Loosening more suture                           \\ \hline
		G11        & Dropping suture at end and moving to end points \\ \hline
		G12        & Reaching for needle with left hand              \\ \hline
		G13        & Making C loop around right hand                 \\ \hline
		G14        & Reaching for suture with right hand             \\ \hline
		G15        & Pulling suture with both hands                  \\ \hline
	\end{tabular}
\end{table}

\vspace{-5pt}
\section{Conclusion}
\vspace{-3pt}
In this paper we propose to use holistic features like SMT, DCT, DFT and ApEn for skill assessment in RMIS training. Our proposed framework out-performs all existing HMM based approaches. We also present a detailed analysis of skill assessment on the JIGSAWS dataset and propose a weighted feature combination technique that further improves performance on score predictions. We do not use any video data making our method computationally feasible for real time feedback. Our framework can easily be integrated in a robotic surgery platform (like the daVinci system) to generate automated modified-OSATS based score reports in training. Moreover, our proposed task highlights generation method could be extremely valuable for giving surgeons more focused feedback.

\bibliographystyle{splncs}
\bibliography{IPCAI_2018}   
\end{document}